\pdfoutput=1
\documentclass[11pt]{article}
\usepackage[preprint]{acl}
\usepackage{times}
\usepackage{latexsym}
\usepackage[T1]{fontenc}
\usepackage[utf8]{inputenc}
\usepackage{microtype}
\usepackage{inconsolata}
\usepackage{graphicx}
\usepackage{kotex}
\usepackage{blindtext}
\usepackage{lipsum}
\usepackage{amssymb}  
\usepackage{multirow}
\usepackage{booktabs} 
\usepackage{tikz}
\usepackage{amssymb}
\usepackage{amsmath}
\usepackage{cleveref}
\usepackage{algorithm}
\usepackage{algpseudocode}
\usepackage{longtable}
\usepackage{subcaption}
\usepackage{tikz}

\newcommand{\methodname}{\textsc{MT-Mol}}
\newlength{\RoundedBoxWidth}
\newsavebox{\GrayRoundedBox}
\newenvironment{GrayBox}[1][\dimexpr\textwidth-4.5ex]%
   {\setlength{\RoundedBoxWidth}{\dimexpr#1}
    \begin{lrbox}{\GrayRoundedBox}
       \begin{minipage}{\RoundedBoxWidth}}%
   {   \end{minipage}
    \end{lrbox}
    \begin{center}
    \begin{tikzpicture}
       \draw node[draw=black,fill=black!10,rounded corners,%
             inner sep=2ex,text width=\RoundedBoxWidth]%
             {\usebox{\GrayRoundedBox}};
    \end{tikzpicture}
    \end{center}}
\title{\methodname: Multi Agent System with Tool-based Reasoning \\for Molecular Optimization}

\author{Hyomin Kim\\
    KAIST \\
  \texttt{hyomin126@kaist.ac.kr} \\\And
  Yunhui Jang \\
  KAIST \\
  \texttt{yunhuijang@kaist.ac.kr} \\\And
  Sungsoo Ahn \\
  KAIST \\
  \texttt{sungsoo.ahn@kaist.ac.kr} \\}

\begin{document}
\maketitle

\begin{abstract}
Large language models (LLMs) have large potential for molecular optimization, as they can gather external chemistry tools and enable collaborative interactions to iteratively refine molecular candidates. However, this potential remains underexplored, particularly in the context of structured reasoning, interpretability, and comprehensive tool-grounded molecular optimization. To address this gap, we introduce \methodname, a multi-agent framework for molecular optimization that leverages tool-guided reasoning and role-specialized LLM agents. Our system incorporates comprehensive RDKit tools, categorized into five distinct domains: structural descriptors, electronic and topological features, fragment-based functional groups, molecular representations, and miscellaneous chemical properties. Each category is managed by an expert analyst agent, responsible for extracting task-relevant tools and enabling interpretable, chemically grounded feedback. \methodname\ produces molecules with tool-aligned and stepwise reasoning through the interaction between the analyst agents, a molecule-generating scientist, a reasoning-output verifier, and a reviewer agent. As a result, we show that our framework shows the state-of-the-art performance of the PMO-1K benchmark on 17 out of 23 tasks. 
\end{abstract}

\section{Introduction}

\begin{figure*}[t]
\centering
    \includegraphics[width=\textwidth]{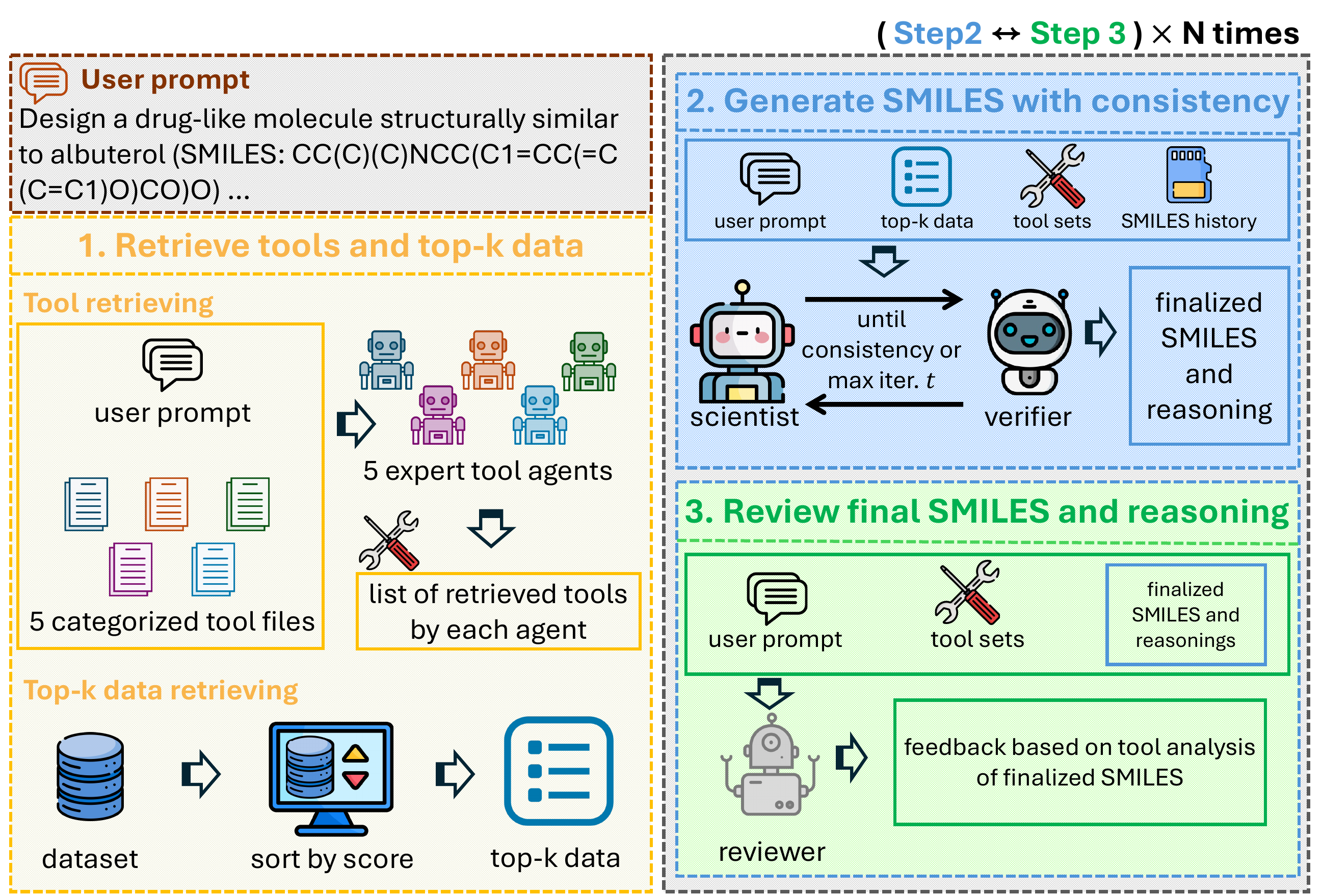}
    \vspace{-0.2in}
    \caption{\textbf{Overview of our method.} Given a molecular optimization task, analyst agents analyze the prompt and outputs list of relevant RDKit functions from five categories. Top-$k$ molecules are retrieved as reference molecules for the scientist agent. Then, the scientist agent proposes a SMILES with stepwise reasoning, which the double checker validates for consistency. The reviewer finally assesses the reasoning using tool-informed descriptors and provides structured feedback. This generation and review process is repeated until the maximum number of iterations $N$ is reached. This multi-agent pipeline enables interpretable, tool-guided molecule generation with iterative refinement toward the design objective.}
    \vspace{-0.2in}
    \label{fig:overview}
\end{figure*}

Large language models (LLMs) have demonstrated remarkable capabilities in a wide range of problems such as question answering \citep{dong2024cost, sun2024harnessing}, summarization \citep{kim2024sure, liu2305learning}, translation \citep{alves2024tower, bari2024allam}, and code generation \citep{chen2021evaluating, li2023starcoder} using large-scale pretraining and in-context learning \citep{brown2020language, chowdhery2023palm, zhang2022opt} \citep{chen2021evaluating}. Motivated by the success, researchers are investigating the potential of LLMs for scientific discovery in the chemical domain \citep{wang2023scientific, luu2020explaining, wang2024molleo, nguyen2024lico, bran2023chemcrow}. 

In particular, employing LLMs to design new molecules (e.g., drug candidates), is promising due to several advantages: (1) LLMs exhibit general understanding and reasoning capabilities obtained from large-scale pretraining, (2) they can use the off-the-shelf tools for analyzing molecules, and (3) they are capable of interact with other agents to further improve the design candidate. 

Recent studies have explored the application of LLMs in molecular optimization. For example, LICO \citep{nguyen2024lico} extends LLMs with embedding layers and in-context examples to build a surrogate modeling framework for molecular optimization. MOLLEO \citep{wang2024molleo} leverages LLMs as mutation and crossover operators within an evolutionary algorithm. ChemCrow~\citep{bran2023chemcrow} integrates LLMs with chemical tools for to faciliate synthesis planning and molecular analysis. While these approaches demonstrate encouraging results, we argue that they do not fully exploit the broader capabilities of modern LLMs such as multi-agent collaboration, tool integration, and iterative reasoning, which are essential for high-quality molecular optimization.

\paragraph{Contribution.} In this work, we propose \methodname{}, a multi-agent framework for molecular optimization. Our key idea is to decompose the optimization process into four distinct roles (\textbf{analyst}, \textbf{scientist},  \textbf{verifier}, and \textbf{reviewer}) and employ specialized agent for each role. Unlike previous approaches, \methodname{} generates molecules with explicit stepwise reasoning, consistency checks, and tool-informed feedback. Furthermore, we collect a set of 154 chemistry-related functions, which serve as applicable tools for agents during molecular generation process.

To be specific, we introduce four agents: (1) analyst, (2) scientist, (3) verifier, and (4) reviewer. In detail, five \textbf{analyst agents} proposes the task-specific relevant tools using different types of chemical functions: structure, electronic properties, functional groups, identifiers, and miscellaneous descriptors. Then a \textbf{scientist agent} proposes new molecules in SMILES format~\citep{weininger1988smiles} and explains each design step through structured reasoning. Next, a \textbf{verifier agent} evaluates whether the reasoning of the scientist is consistent with the proposed molecule. Finally, a \textbf{reviewer agent} assesses both the molecule and the reasoning process using the outputs from the tools and provides detailed feedback. Each agent plays a collaborative role that enables interpretable, tool-aware, and iterative molecular design. By incorporating domain-specific tools such as RDKit \citep{landrum2013rdkit}, \methodname\ supports chemically informed generation and transparent decision-making. 

In summary, we propose a multi-agent framework for molecular optimization, where each agent is assigned a specific role such as tool selection, molecule generation, consistency validation, and reasoning critique. Our system integrates 154 RDKit functions, organized into five specialized analyst agents covering structural descriptors, electronic and topological descriptors, structural descriptors, fragment-based analysis, and identifiers or representations. We achieve state-of-the-art performance on 17 out of 23 tasks from the PMO-1K benchmark, outperforming recent strong baselines including LICO and MOLLEO in terms of top-10 AUC scores. Additionally, our framework offers an interpretable reasoning pipeline in which each generated molecule is equipped with stepwise rationale, double-check verification, and tool-informed reviewer feedback.
\section{Related Work}

\paragraph{Generative models for molecular optimization.} Molecular optimization aims to design molecules that maximize desired chemical or biological properties, such as solubility, binding affinity, or synthesizability. Generative modeling has emerged as a central approach for this task, encompassing techniques from deep learning to probabilistic search. REINVENT \citep{olivecrona2017reinvent} introduced reinforcement learning over SMILES strings to fine-tune molecular generation toward desired properties. \citet{jensen2019graphga} showed that graph-based genetic algorithms and non-ML models combined with Monte Carlo Tree Search perform competitively in optimizing molecular properties under synthetic constraints. Augmented Memory \citep{guo2024augmented} enhances sample efficiency in reinforcement learning through SMILES augmentation and experience replay. Genetic GFN \citep{bengio2023gflownet} enables compositional molecule generation by sampling in proportion to a reward function, offering diversity and high-reward sampling in molecular benchmarks. \citet{srinivas2009gpbo} introduced GP BO, a Gaussian process-based optimization framework that provides sublinear regret bounds and sample-efficient exploration using information gain from kernel-based uncertainty modeling. While these models improve sample efficiency and diversity, they often lack interpretability and fail to fully utilize the available domain knowledge, such as chemical priors. Our framework complements these approaches by incorporating structured reasoning and chemical tools into the molecular generation process.

\paragraph{LLMs for molecular optimization.} LLMs have recently been applied to molecular optimization tasks. LICO \citep{nguyen2024lico} extends a pretrained LLM with structured embeddings to model property functions without relying on natural language prompts. MOLLEO \citep{wang2024molleo} uses LLMs as evolutionary operators, enabling coherent molecule generation across single- and multi-objective settings. Prompt-MolOpt \citep{wu2024promptmolopt} introduces prompt-based editing to optimize multiple properties in low-data regimes while preserving pharmacophores. DrugAssist \citep{ye2025drugassist} fine-tunes an instruction-based LLM on a curated chemistry dataset to support interactive, feedback-driven molecule design. ChemCrow \citep{bran2023chemcrow} combines general-purpose LLMs with chemistry tools and a ReAct-based reasoning loop to automate generation, retrosynthesis, and property prediction. Despite these advances, existing approaches often lack interpretability, structured collaboration among specialized agents, and a systematic feedback loop that enhances accurate molecule design. To address these limitations, our method introduces five expert analyst agents powered by RDKit~\citep{landrum2013rdkit} and a multi-agent feedback loop that ensures both accurate and interpretable molecular optimization.

\paragraph{Multi-agent LLMs.} Multi-agent LLMs have shown promise in collaborative reasoning and decomposed problem-solving. AgentVerse \citep{chen2023agentverse} assigns agents to roles like recruitment and evaluation, leveraging specialization for better coordination. ProAgent \citep{zhang2024proagent} enables agents to infer and adapt to teammates’ strategies through communication history. Self-Adaptive Multi-agent Systems \citep{nascimento2023self} use a self-control loop to make agents responsive to dynamic environments. Theory of Mind for Multi-Agent Collaboration \citep{li2023theory} enhances coordination by giving agents shared belief states and goal-tracking abilities. MetaGPT \citep{hong2023metagpt} improves communication scalability via a Shared Message Pool that standardizes agent interactions. While these frameworks contribute to multi-agent architectural design, they have overlooked domain-specific tool integration and have less focused on molecule optimization. Our framework addresses this gap by tightly coupling expert analyst agents with reasoning roles to enable targeted, tool-informed molecular design.

\section{Method}
In this section, we introduce our multi-agent framework for molecular optimization, coined \methodname{}. In \Cref{sec:overall_framework}, we first describe the overview of our system, which consists of four primary agent types: 1) analyst, 2) scientist, 3) verifier, and 4) reviewer agents. We describe details of the analysts in \Cref{sec:details_of_tool_agents}, and stepwise reasoning and feedback process in \Cref{sec:reasoning_feebdack}.

\begin{figure*}[htp!]
\centering
    \includegraphics[height=9.2cm,width=\textwidth]{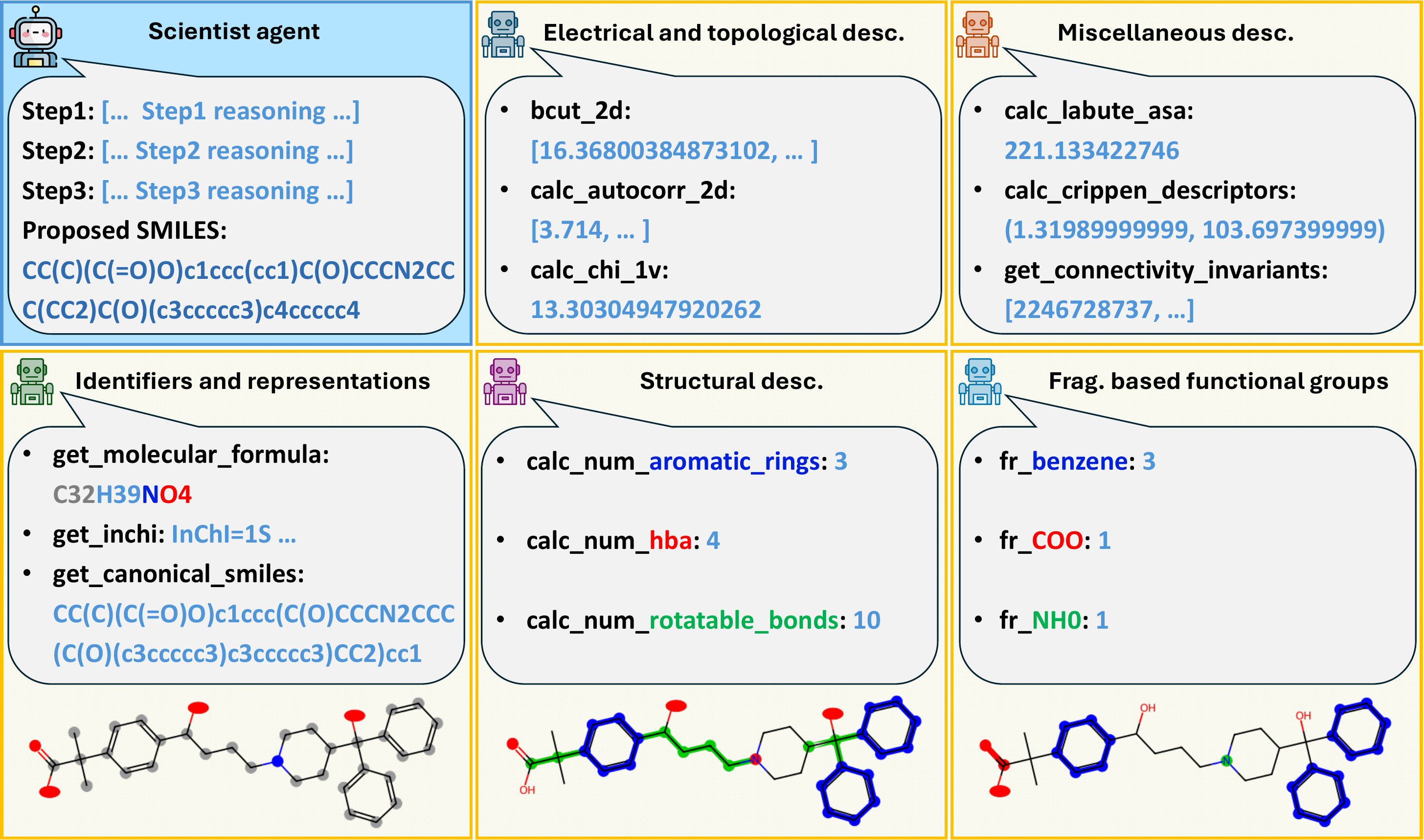}
    \caption{\textbf{Example of analyst agents.} Example case of five analyst agents analyzing the SMILES proposed by the scientist agent for the fexofenadine\_mpo task. Each analyst agent chooses task-relevant tools: electronical and topological descriptors, miscellaneous descriptors, identifiers and representations, structural descriptors, and functional groups. The molecules at the bottom visualizes how analyst agents analyze the scientist agent's proposed SMILES. We provide the description of the tools at  \Cref{appdx:tool_list}.}
    \label{fig:tool_usage}
    \vspace{-0.1in}
\end{figure*}
\subsection{Overall Framework} \label{sec:overall_framework}
In this section, we present a high-level overview of our multi-agent framework for molecular optimization. Given a user prompt $T$, analyst agents first select relevant tools, then the scientist agent proposes a molecule with structured reasoning. The verifier agent then verifies the logical consistency of the proposed output. Finally, the reviewer agent provides detailed feedback grounded in chemical analysis tools. We provide an overview of our method in \Cref{fig:overview}. 

\begin{figure*}[t]
    \centering

    \begin{subfigure}[b]{\textwidth}
        \centering
        \includegraphics[width=\textwidth]{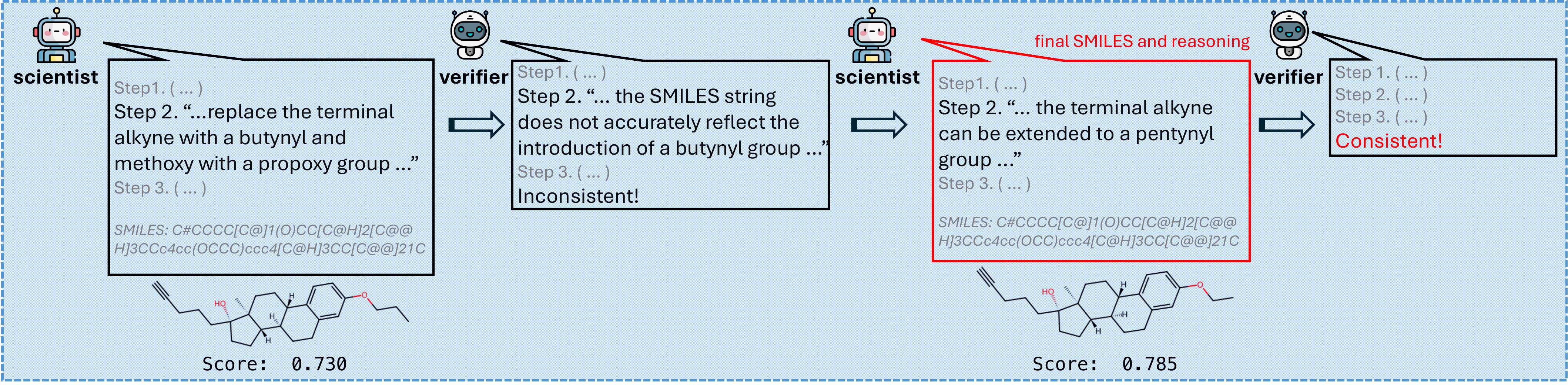}
        \caption{Structured feedback of the verifier agent.}
        \label{fig:st_double_checker}
    \end{subfigure}

    \vspace{0.5cm} 

    \begin{subfigure}[b]{0.85\textwidth}
        \centering
        \includegraphics[width=\textwidth]{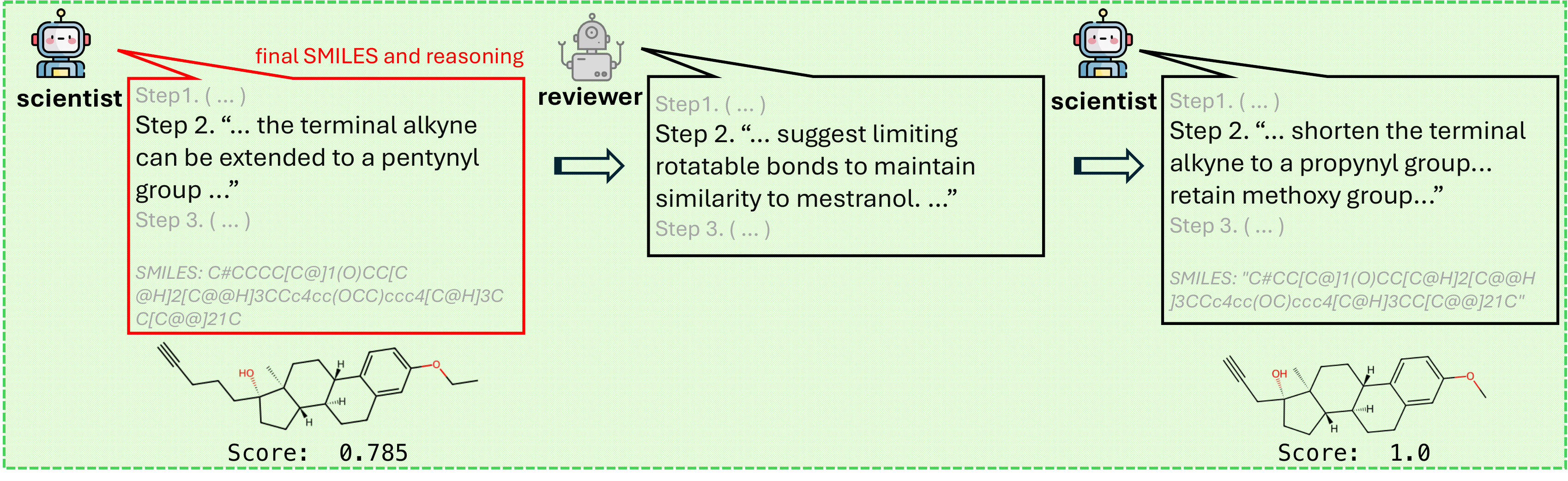}
        \caption{Structured feedback of the reviewer agent.}
        \label{fig:st_reviewer}
    \end{subfigure}

    \caption{\textbf{Examples of structured and stepwise response.} The figures illustrate examples of structured feedback mechanisms employed by our agent system for the mestranol\_similarity task. (a) The verifier flags a mismatch between reasoning and SMILES and the scientist revises both for consistency. (b) The reviewer suggests reducing rotatable bonds and the scientist reflects the design, improving the score.}
\vspace{-0.2in}
    \label{fig:structured_feedback}
\end{figure*}

Notably, our agents are informed about the details of the objective function and utilize their chemical knowledge to propose better molecules. This is in contrast to existing non-LLM works in molecular design that assume black-box objective functions. We believe that this is a strength of our approach, since in most of the tasks, we have some information about the objective function that can be described in natural language.
\vspace{-0.2in}
\paragraph{Analysts.} We design five analyst agents for different aspects of molecular analysis. Each analyst agent parses and analyzes the molecule in the task prompt $T$ and the scientist agent's proposed SMILES. Each analyst agent wraps a curated set of RDKit or PubChem functions in one of the following categories: 1) structural descriptors, 2) electronic and topological descriptors, 3) fragment-based functional group detectors, 4) chemical identifiers and representations, and 5) miscellaneous descriptors agents.
To analyze a task prompt $T$, the analyst agents identify the most relevant chemical features and select the tools accordingly. We illustrate the example case of how tools are used in \Cref{fig:tool_usage} and provide the details of the tool in \Cref{appdx:tool_list}.

\paragraph{Scientist.} The scientist agent generates a novel molecule in SMILES format, denoted $S$, along with a reasoning path for proposing the molecule. To this end, the agent utilizes the tool-based analysis of the task prompt and a history of previously generated SMILES to avoid duplication.
Based on the collected information, the agent proposes a molecule design strategy. It outlines this strategy in a sequence of $k$ reasoning steps $\lbrace r_1,\ldots,r_k\rbrace$, where each $r_i$ explains how the scientist agent thinks when proposing the SMILES representation of a molecule. After the reasoning process, the agent generates a SMILES string $S$.

\paragraph{Verifier.} As noted by \citet{pan2025multiagent}, reasoning–action mismatch is a critical issue in multi-agent frameworks. To mitigate this in our system, we introduce a verifier agent that verifies each reasoning step in $\lbrace r_1,\ldots,r_k\rbrace$, ensuring that every $r_i$ is faithfully reflected in the proposed SMILES $S$. In detail, it parses each step $r_i$ and examines whether $S$ contains the corresponding molecular feature. When discrepancies arise (e.g., when a reasoning step claims the presence of a nitro group, but $S$ lacks it), the agent flags the inconsistency and produces stepwise feedbacks $\lbrace f^v_1,\ldots, f^v_k\rbrace$. Then, the verifier asks the scientist to re-generate the SMILES based on the feedback. This re-generation loop continues until the verifier confirms consistency between the reasoning and SMILES, or until a maximum number of iterations $t$ reached. If there is no discrepancy detected, it passes the verified reasoning steps $\lbrace r_1,\ldots,r_k\rbrace$ and SMILES $S$ to the reviewer agent.

\vspace{-2mm}
\paragraph{Reviewer.} Inspired by previous works using LLMs as reviewers \citep{hosseini2023fighting, zhang2022investigating}, we introduce a chemical reviewer agent that evaluates and provides informative feedback. Specifically, the reviewer agent evaluates the verified SMILES $S$ and reasoning steps $\lbrace r_1,\ldots,r_k\rbrace$. Using tool-based analysis of $S$, it provides chemically grounded, stepwise feedback $\lbrace f^r_1,\ldots,f^r_k\rbrace$ aligned with the structure of the reasoning. This feedback includes confirmations of correct reasoning, identification of wrong or missing claims, and suggestions for revision. The scientist agent then uses this feedback to refine both the reasoning and molecule $S$ in the next iteration, enabling iterative improvement.

\subsection{Details of analyst agents}\label{sec:details_of_tool_agents}
\vspace{-2mm}
We implement our multi-agent system with specialized LLM agents, with analyst agents playing a key role in analyzing molecules using domain-specific RDKit \citep{landrum2013rdkit} functions. These tools guide molecule generation by providing relevant descriptors to the scientist agent and support the reviewer with interpretable feedback. To enable comprehensive tool utilization and decomposed analysis, we categorize the analyst agents into five molecule-specialized aspects. Each agent targets a distinct aspect of molecular analysis and contributing to a chemically informed and interpretable design process. We provide a detailed list of tools that analyst agents take at \Cref{appdx:tool_list}.
\begin{table*}[hpt!]
\centering
\resizebox{\textwidth}{!}{
\begin{tabular}{lccccccccccc}
\toprule
\textbf{Task} & \textbf{GP BO} & \textbf{REINVENT} & \textbf{LICO}\textit{-L} & \textbf{Genetic GFN} & \textbf{Graph GA} & \textbf{Aug. Mem.} & \textbf{MOLLEO}\textit{-B} & \textbf{MOLLEO}\textit{-D}* & \textbf{\methodname}\textit{-D}* \\
\midrule
albuterol\_similarity & $0.636$ & $0.496$ & $0.656$ & $0.664$ & $0.583$ & $0.557$ & $\underline{{0.886}}$ & $0.883$ & ${\mathbf{0.998}}$ \\
amlodipine\_mpo & $0.519$ & $0.472$ & $0.541$ & $0.534$ & $0.501$ & $0.489$ & $\underline{{0.637}}$ & $0.540$ & ${\mathbf{0.647}}$ \\
celecoxib\_rediscovery & $0.411$ & $0.370$ & $0.447$ & $0.447$ & $0.424$ & $0.385$ & $0.402$ & $\underline{{0.512}}$ & ${\mathbf{0.867}}$ \\
deco\_hop & $0.593$ & $0.572$ & $0.596$ & $\underline{{0.604}}$ & $0.581$ & $0.579$ & $0.588$ & $0.574$ & ${\mathbf{0.842}}$ \\
drd2 & $0.857$ & $0.775$ & $\underline{{0.859}}$ & $0.809$ & $0.833$ & $0.795$ & ${\mathbf{0.910}}$ & $0.812$ & $0.756$ \\
fexofenadine\_mpo & $\underline{{0.707}}$ & $0.650$ & $0.700$ & $0.682$ & $0.666$ & $0.679$ & $0.674$ & $0.680$ & ${\mathbf{0.883}}$ \\
gsk3b & $0.611$ & $0.589$ & $\underline{{0.617}}$ & ${\mathbf{0.637}}$ & $0.523$ & $0.539$ & $0.397$ & $0.496$ & $0.308$ \\
isomers\_c7h8n2o2 & $0.545$ & $0.725$ & $0.779$ & $0.738$ & $0.735$ & $0.661$ & $0.737$ & $\underline{{0.850}}$ & ${\mathbf{0.986}}$ \\
isomers\_c9h10n2o2pf2cl & $0.599$ & $0.630$ & $0.672$ & $0.656$ & $0.630$ & $0.596$ & $0.635$ & $\underline{{0.832}}$ & ${\mathbf{0.914}}$ \\
jnk3 & $\underline{{0.346}}$ & $0.315$ & $0.336$ & ${\mathbf{0.409}}$ & $0.301$ & $0.294$ & $0.186$ & $0.342$ & $0.125$ \\
median1 & $0.213$ & $0.205$ & $0.217$ & $0.219$ & $0.208$ & $0.219$ & $\underline{{0.236}}$ & $0.193$ & ${\mathbf{0.321}}$ \\
median2 & $0.203$ & $0.188$ & $0.193$ & $\underline{{0.204}}$ & $0.181$ & $0.184$ & $0.191$ & $0.197$ & ${\mathbf{0.322}}$ \\
mestranol\_similarity & $0.427$ & $0.379$ & $0.423$ & $0.414$ & $0.362$ & $0.393$ & $0.399$ & $\underline{{0.630}}$ & ${\mathbf{0.996}}$ \\
osimertinib\_mpo & $0.766$ & $0.737$ & $0.759$ & $0.763$ & $0.751$ & $0.761$ & $\underline{{0.779}}$ & $0.753$ & ${\mathbf{0.796}}$ \\
perindopril\_mpo & $0.458$ & $0.404$ & $0.473$ & $0.462$ & $0.435$ & $0.422$ & ${\mathbf{0.655}}$ & $0.422$ & $\underline{{0.542}}$ \\
qed & $0.912$ & $0.921$ & $\underline{{0.925}}$ & ${\mathbf{0.928}}$ & $0.914$ & $0.923$ & $0.919$ & $\mathbf{0.928}$ & $0.903$ \\
ranolazine\_mpo & $\mathbf{0.701}$ & $0.574$ & $\underline{{0.687}}$ & $0.623$ & $0.620$ & $0.614$ & $0.640$ & $0.516$ & $0.233$ \\
scaffold\_hop & $0.478$ & $0.447$ & $0.480$ & $\underline{{0.485}}$ & $0.461$ & $0.460$ & $0.473$ & $0.464$ & ${\mathbf{0.646}}$ \\
sitagliptin\_mpo & $0.232$ & $0.261$ & $\underline{{0.315}}$ & $0.227$ & $0.229$ & $0.245$ & $0.193$ & ${\mathbf{0.328}}$ & $0.067$ \\
thiothixene\_rediscovery & $0.351$ & $0.311$ & $0.343$ & $0.377$ & $0.322$ & $0.336$ & $0.416$ & $\underline{{0.478}}$ & ${\mathbf{0.719}}$ \\
troglitazone\_rediscovery & $0.313$ & $0.246$ & $0.292$ & $0.277$ & $0.267$ & $0.262$ & $0.302$ & $\underline{{0.387}}$ & ${\mathbf{0.841}}$ \\
valsartan\_smarts & $0.000$ & $0.000$ & $0.000$ & $0.000$ & $0.000$ & $0.000$ & $0.000$ & $0.000$ & $0.000$ \\
zaleplon\_mpo & $0.392$ & $0.406$ & $0.404$ & $0.400$ & $0.374$ & $\underline{{0.415}}$ & $0.392$ & $0.409$ & ${\mathbf{0.625}}$ \\
\midrule
\textbf{Sum of scores (↑)} & $11.27$ & $10.68$ & $11.71$ & $11.56$ & $10.90$ & $10.81$ & $11.65$ & $\underline{{12.23}}$ & ${\mathbf{15.42}}$ \\
\bottomrule
\end{tabular}
}
\caption{\textbf{Results of PMO-1K benchmark.} Tasks are assessed using AUC top-10 averaged by multiple runs. Results with (*) are evaluated from 3 independent runs while the others are assessed from 5 independent runs. We mark the best result in {\textbf{bold}} and the second-best in \underline{underline} for each task.}
\vspace{-0.2in}
\label{tab:main_pmo_table}
\end{table*}

\paragraph{Electronic and topological descriptors.} This agent analyzes how electrons are distributed in a molecule and how its atoms are connected, helping to assess properties such as reactivity and stability. It captures patterns that are important for determining whether a molecule is likely to behave well as a drug. As shown in \Cref{fig:tool_usage}, this includes features such as charge distribution.

\paragraph{Fragment-based functional groups.} This agent breaks molecules down into recognizable building blocks, such as rings or functional groups, such as acids or amines, which are commonly used in chemistry. These fragments are easy to interpret and often appear in the stepwise reasoning provided by the scientist agent. \Cref{fig:tool_usage} shows how the agent highlights specific substructures, such as aromatic rings, that are captured, which is a key component of the task.

\paragraph{Identifiers and representations.} This agent translates molecules into standardized formats such as canonicalized SMILES representation, molecular formulas, etc. \Cref{fig:tool_usage} illustrates how the functional group agent identifies chemically significant motifs such as benzene rings and carboxylic acids, which reflect specific fragment-level reasoning steps and enable chemically grounded feedback.

\paragraph{Structural descriptors.} This agent captures basic geometric and physical features of a molecule, such as the number of atoms or bonds it contains. These properties influence how a molecule might behave in real-world conditions, including how it binds to targets or dissolves. As shown in \Cref{fig:tool_usage}, this agent helps evaluate aspects like bond rotatability or ring complexity.

\paragraph{Miscellaneous descriptors.} Miscellaneous descriptors agent provides additional analysis that complements the outputs of other agents. It captures properties that might be overlooked, such as molecular surface area, hybridization patterns, or structural irregularities, and helps ensure that the generated molecule is chemically reasonable. As shown in \Cref{fig:tool_usage}, it offers supplementary evidence that strengthens the overall reasoning process.

\subsection{Structured and stepwise response}\label{sec:reasoning_feebdack} 
In order to ensure that the agent responds to every desired component (e.g., stepwise reasoning, feedback, and SMILES), we guide the agent to output in JSON format using OpenAI API's function\footnote{\url{https://platform.openai.com/docs/guides/structured-outputs?api-mode=chat}}. Specifically, the scientist agent generates stepwise reasoning and SMILES, while the verifier and reviewer agents produce stepwise feedback in a designated JSON format.

Also, for a valid and interpretable response, we guide the agents to output stepwise reasoning and feedback. Specifically, the scientist output stepwise reasoning $\lbrace r_1,\ldots, r_k\rbrace$ when proposing a SMILES~$S$. Then, the verifier agent ensures the scientist agent's stepwise reasoning $\lbrace r_1,\ldots,r_k\rbrace$ is consistent with the output SMILES by providing the interpretable feedback $\lbrace f^d_1,\ldots, f^d_k\rbrace$. We visualize the example case in \Cref{fig:st_double_checker}. The verifier agent identifies an inconsistency between the scientist’s reasoning and the SMILES, since the butynyl group is not encoded. 

In addition, the reviewer critiques the reasoning of the scientist agent with stepwise feedback $\lbrace f^r_1,\ldots, f^r_k\rbrace$. As illustrated in \Cref{fig:st_reviewer}, the reviewer agent highlights the issue of increased rotatable bonds. This leads the scientist to revise the design by shortening the alkyne and restoring the methoxy group, which significantly improves the structural similarity score to 1.0. This shows that our approach enables high alignment to target molecule, interpretability, and validity of the properties. We provide a detailed prompt and response example in \cref{appdx:prompts}.

\begin{figure*}[t]
\centering
    \includegraphics[width=\textwidth]{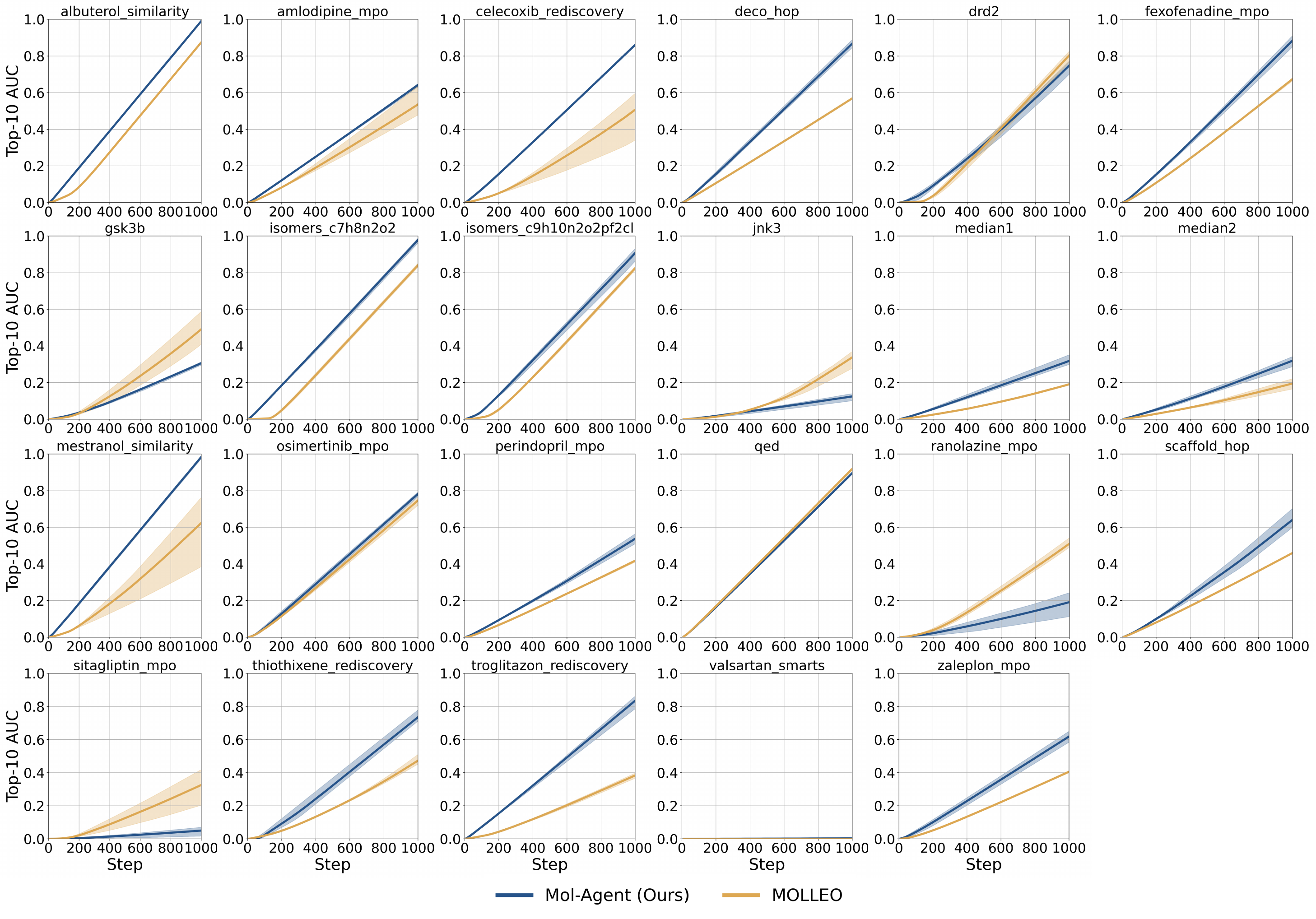}
    \caption{\textbf{Top-10 AUC curves.} Top-10 average AUC curves on the PMO benchmark, averaged over three random seeds. Our method consistently surpasses MOLLEO by achieving higher and faster-rising AUC curves, highlighting the effectiveness of tool-guided reasoning and multi-agent feedback in molecular optimization.}
    \vspace{-0.1in}
    \label{fig:auc_graph}
\end{figure*}

\section{Experiments}
\vspace{-0.5mm}
In this section, we evaluate the effectiveness of our multi-agent LLM system for molecular optimization in low-budget settings. We conduct experiments on the practical molecular optimization (PMO)-1K benchmark, which contains 23 chemically diverse optimization tasks, ranging from rediscovery and scaffold hopping to multi-property objectives. Our framework consists of expert analyst agents—each specialized in task decomposition, SMILES generation, verification, and tool-informed feedback—that collaborate to produce interpretable and high-quality molecular optimization. We compare our results against existing LLM-driven and evolutionary baselines, including LICO and MOLLEO, using various backbone models. We describe the dataset and baselines below, followed by the experimental setting described in \Cref{subsec:pmo}. We then present the main benchmark results in Table 1 and provide analysis in \Cref{subsec:ablation}.
\vspace{-0.1in}
\paragraph{Datasets.} 
We evaluate on the Practical Molecular Optimization benchmark \citep{gao2022sample}, which comprises 23 molecular optimization tasks. Each task defines a specific molecular property or structural constraint, such as rediscovery of known drugs (e.g., celecoxib, thiothixene), similarity to target scaffolds, or maximization of molecular property scores such as quantitative estimate of drug-likeness (QED) or logP. Following \citet{gao2022sample}, we assess performance using the top-10 area under the curve (AUC), which measures the average property score over oracle calls. Additionally considering \citet{nguyen2024lico}, the evaluation is conducted for 1K oracle calls, simulating a budget-constrained discovery setting. We use the ZINC 250K \citep{sterling2015zinc} dataset to retrieve the top-100 reference molecules for the scientist agent’s prompt. We summarize the entire tasks and their descriptions in \Cref{appdx:task_description}. All molecules are represented in the SMILES format and evaluated using predefined black-box scoring functions consistent with the PMO benchmark protocol.
\vspace{-2mm}
\paragraph{Baselines.} %TODO: Add citations
We compare our framework against six baselines: GP BO \citep{srinivas2009gpbo}, REINVENT \citep{olivecrona2017reinvent}, LICO \citep{nguyen2024lico}, and two variants of MOLLEO \citep{wang2024molleo} (MOLLEO-B, and MOLLEO-D). MOLLEO operates through LLM-guided mutation and crossover, using different base models (BioT5 \citep{pei2023biot5} and DeepSeek-V3 \citep{liu2024deepseek}. We evaluated two versions of our framework (Ours-$D$) using DeepSeek-V3 as a backbone for all the agent roles.

\vspace{-0.1in}
\subsection{PMO Benchmark}\label{subsec:pmo}

\Cref{tab:main_pmo_table} reports the performance of our framework and competing methods in all 23 PMO tasks. \methodname \text{-}D*, achieves the best performance in 17 of 23 tasks, significantly outperforming all baselines, including MOLLEO and LICO. In particular, \methodname\ surpasses the SOTA AUC sum of 12.23 (MOLLEO-D*) with a score of 15.42, marking a substantial improvement in the overall efficiency of optimization. The performance gap is particularly large on chemically complex tasks such as celecoxib\_rediscovery and amlodipine\_mpo, where \methodname\text{-}D* outperforms the previous best by more than 0.3 AUC points.

In \Cref{fig:auc_graph}, we visualize the top-10 AUC curves for every 23 PMO tasks. \methodname\ consistently achieves faster and higher AUC trajectories compared to MOLLEO-D* across tasks such as albuterol\_similarity, amlodipine\_mpo, osimertinib\_mpo, and troglitazon\_rediscovery. These results suggest that our tool-aware reasoning, stepwise validation, and multi-agent feedback loop generate the desired molecule SMILES in the early stage while achieving high oracle value. The improvements are especially pronounced in the early stages of generation, indicating that Mol-Agent makes more efficient use of oracle calls.

\begin{table}[t]
\centering
\resizebox{\columnwidth}{!}{
    \begin{tabular}{ll|c}
    \toprule
    \textbf{Task} & \textbf{Setting} & \textbf{AUC-Top10} \\
    \midrule
    \multirow{4}{*}{osimertinib\_mpo} 
        & \methodname        & $\mathbf{0.796\pm0.005}$\\
        & w/o Tool           & $0.694\pm0.054$\\ 
        & w/o Reviewer       & $0.619\pm0.140$\\ 
        & w/o Double checker & $0.704\pm0.017$\\ 
    \midrule
    \multirow{4}{*}{albuterol\_similarity} 
        & \methodname        & $\mathbf{0.998\pm0.000}$\\
        & w/o Tool           & $0.750\pm0.021$\\ 
        & w/o Reviewer       & $0.991\pm0.003$\\ 
        & w/o Double checker & $0.996\pm0.003$\\ 
    \midrule
    \multirow{4}{*}{mestranol\_similarity} 
        & \methodname        & $\mathbf{0.996\pm0.001}$\\
        & w/o Tool           & $0.831\pm0.052$\\ 
        & w/o Reviewer       & $0.990\pm 0.002$ \\ 
        & w/o Double checker & $0.994\pm0.002$\\ 
    \bottomrule
    \end{tabular}
}
\caption{\textbf{Ablation study.} AUC-Top10 score under different agent removals for each task.}
\label{tab:ablation}
\end{table}
\vspace{-0.05in}

\subsection{Ablation studies}\label{subsec:ablation}
To evaluate the contribution of each component in our multi-agent framework, we perform an ablation study on a subset of tasks from the PMO benchmark. Specifically, we assess the impact of removing (1) all five expert analyst agents, (2) the verifier agent, and (3) the reviewer agent. We report top-10 AUC scores averaged over three random seeds in \Cref{tab:ablation}.

One can observe that removing the analyst agents consistently leads to a substantial drop in performance across all tasks. For instance, the AUC score on albuterol\_similarity drops from 0.998 to 0.750, highlighting that the expert analyst agents provide essential domain-specific descriptors.

Also removing reviwer agent causes noticeable degradation on tasks like osimertinib\_mpo (from 0.796 to 0.619). Similarly, removing verifier aent shows modest performance drops when ablated, particularly on more challenging tasks.

Overall, these results underscore the importance of tool-guided analysis, structured reasoning verification, and feedback loops in our multi-agent system.
\section{Conclusion}
In this paper, we introduced \methodname, a multi-agent framework for molecular optimization and generation that combines tool-guided reasoning with structured collaboration among specialized LLM agents.  Our system integrates five expert analyst agents, each equipped with domain-specific chemistry functions, to guide and critique molecule design. Through systematic interaction among scientist, verifier, and reviewer agents, \methodname\ achieves interpretable, chemically valid, and task-aligned molecular optimization. Our experiments on the PMO-1K benchmark demonstrate that \methodname\ outperforms strong baselines, including LICO and MOLLEO, achieving state-of-the-art performance on 17 out of 23 tasks. The results highlight the effectiveness of structured reasoning, tool-based validation, and multi-agent feedback in navigating the complex chemical space. This work provides the multi-agent system with comprehensive and systematic tool-augmented responses, accelerating molecular optimization and enabling transparent scientific discovery.

\clearpage
\section*{Broader Impact}
Our work may help democratize access to molecular design expertise by enabling non-expert users to interact with intelligent agents that provide chemically grounded suggestions. Furthermore, the stepwise reasoning and feedback mechanisms embedded in our framework can serve as educational tools to help students and researchers understand the rationale behind molecule design decisions.

However, broader adoption of AI-assisted molecule design systems also raises potential ethical and social concerns. These include the misuse of generative tools for designing harmful substances, propagation of biases present in pretraining data, and the risk of over-reliance on AI-generated outputs without sufficient domain validation. Responsible deployment will require integrating safety checks, transparency mechanisms, and human-in-the-loop oversight.

\section*{Limitations}
Our framework relies heavily on rule-based cheminformatics tools (e.g., RDKit) and predefined feature sets, which may limit generalization to novel chemical spaces or underrepresented functional groups. Moreover, while the multi-agent structure enables interpretability, it introduces additional computational overhead compared to single-agent models, potentially limiting scalability in resource-constrained settings.

Additionally, our experiments are conducted only in English and do not explore across other languages. This may limit usability in multilingual research environments or for integration with non-English scientific literature and databases.

\section*{Acknowledgments}
This work was partly supported by Institute for Information \& communications Technology Planning \& Evaluation(IITP) grant funded by the Korea government(MSIT) (RS-2019-II190075, Artificial Intelligence Graduate School Support Program(KAIST)), National Research Foundation of Korea(NRF) grant funded by the Ministry of Science and ICT(MSIT) (No. RS-2022-NR072184), GRDC(Global Research Development Center) Cooperative Hub Program through the National Research Foundation of Korea(NRF) grant funded by the Ministry of Science and ICT(MSIT) (No. RS-2024-00436165), and the Institute of Information \& Communications Technology Planning \& Evaluation(IITP) grant funded by the Korea government(MSIT) (RS-2025-02304967, AI Star Fellowship(KAIST)).

\bibliography{custom}
\newpage
\clearpage
\appendix
\newpage
\section{List of tools}\label{appdx:tool_list}
\begin{table*}[ht]
\centering
\begin{tabular}{|l|p{0.58\textwidth}|}
\hline
\textbf{Function} & \textbf{Description} \\
\hline
\texttt{bcut2d} & Implements BCUT descriptors From J. Chem. Inf. Comput. Sci., Vol. 39, No. 1, 1999Diagonal elements are (currently) atomic mass, gasteiger charge,crippen logP and crippen MRReturns the 2D BCUT2D descriptors vector as described in returns [mass eigen value high, mass eigen value low, gasteiger charge eigenvalue high, gasteiger charge low, \\
\hline
\texttt{calcautocorr2d} & Returns 2D Autocorrelation descriptor vector using a specified atom property. \\
\hline
\texttt{calcchi0n} & Calculates the Chi0n index, a valence-based topological descriptor. \\
\hline
\texttt{calcchi0v} & Calculates the Chi0v index, a non-valence-based topological descriptor. \\
\hline
\texttt{calcchi1n} & Calculates the Chi1n index using atom connectivity and optionally forces
calculation. \\
\hline
\texttt{calcchi1v} & Calculates the Chi1v index, a valence-corrected form of Chi1n. \\
\hline
\texttt{calcchi2n} & Calculates the Chi2n index, a higher-order topological descriptor (non-valence-
based). \\
\hline
\texttt{calcchi2v} & Calculates the Chi2n index, a higher-order topological descriptor (non-valence-
based).\\
\hline
\texttt{calcchi3n} & Calculates the Chi3n index for extended connectivity (non-valence-based). \\
\hline
\texttt{calcchi3v} & Calculates the Chi3v index with valence correction for deeper molecular topology. \\
\hline
\texttt{calcchi4n} & Calculates the Chi4n index, further extending non-valence connectivity descriptors. \\
\hline
\texttt{calcchi4v} & Calculates the Chi4v index, a valence-aware descriptor at a 4th topological level \\
\hline
\end{tabular}
\caption{List of Electronic Topological Descriptors tools}
\label{tab:Electronic_Topological_Descriptors}
\end{table*}

\begin{table*}[ht]
\centering
\begin{tabular}{|l|p{0.58\textwidth}|}
\hline
\textbf{Function} & \textbf{Description} \\
\hline
\texttt{get\_min\_ring\_frequency} & Return the least frequent known ring system in the molecule with its frequency. \\
\hline
\texttt{remove\_stereo\_from\_smiles} & Removes stereochemistry from SMILES and returns canonical SMILES and InChI Key. \\
\hline
\texttt{get\_spiro\_atoms} & Returns atom indices that are shared between two rings (spiro atoms). \\
\hline
\texttt{max\_ring\_size} & Returns the size of the largest ring in the molecule. \\
\hline
\texttt{ring\_stats} & Returns the number of rings and the size of the largest ring in the molecule. \\
\hline
\texttt{count\_fragments} & Returns the number of molecular fragments present in the SMILES. \\
\hline
\texttt{get\_largest\_fragment} & Returns the SMILES of the largest fragment by atom count in a molecule. \\
\hline
\texttt{fr\_phos\_acid} & Number of phosphoric acid groups \\
\hline
\texttt{fr\_Al\_COO} & Number of aliphatic carboxylic acids \\
\hline
\texttt{fr\_Al\_OH} & Number of aliphatic hydroxyl groups \\
\hline
\texttt{fr\_Al\_OH\_noTert} & Number of aliphatic hydroxyl groups excluding tert-OH \\
\hline
\texttt{fr\_ArN} & Number of N functional groups attached to aromatics \\
\hline
\texttt{fr\_Ar\_COO} & Number of Aromatic carboxylic acid \\
\hline
\texttt{fr\_Ar\_N} & Number of aromatic nitrogens \\
\hline
\texttt{fr\_Ar\_NH} & Number of aromatic amines \\
\hline
\texttt{fr\_Ar\_OH} & Number of aromatic hydroxyl groups \\
\hline
\texttt{fr\_COO} & Number of carboxylic acids \\
\hline
\texttt{fr\_COO2} & Number of carboxylic acids \\
\hline
\texttt{fr\_C\_O} & Number of carbonyl O \\
\hline
\texttt{fr\_C\_O\_noCOO} & Number of carbonyl O, excluding COOH \\
\hline
\texttt{fr\_C\_S} & Number of thiocarbonyl \\
\hline
\texttt{fr\_HOCCN} & Number of C(OH)CCN-Ctert-alkyl or C(OH)CCNcyclic \\
\hline
\texttt{fr\_Imine} & Number of Imines \\
\hline
\texttt{fr\_NH0} & Number of Tertiary amines \\
\hline
\texttt{fr\_NH1} & Number of Secondary amines \\
\hline
\texttt{fr\_NH2} & Number of Primary amines \\
\hline
\texttt{fr\_N\_O} & Number of hydroxylamine groups \\
\hline
\texttt{fr\_Nhpyrrole} & Number of H-pyrrole nitrogens \\
\hline
\texttt{fr\_SH} & Number of thiol groups \\
\hline
\texttt{fr\_aldehyde} & Number of aldehydes \\
\hline
\texttt{fr\_alkyl\_carbamate} & Number of alkyl carbamates (subject to hydrolysis) \\
\hline
\texttt{fr\_alkyl\_halide} & Number of alkyl halides \\
\hline
\texttt{fr\_allylic\_oxid} & Number of allylic oxidation sites excluding steroid dienone \\
\hline
\texttt{fr\_amide} & Number of amides \\
\hline
\texttt{fr\_amidine} & Number of amidine groups \\
\hline
\texttt{fr\_aniline} & Number of anilines \\
\hline
\texttt{fr\_aryl\_methyl} & Number of aryl methyl sites for hydroxylation \\
\hline
\texttt{fr\_azide} & Number of azide groups \\
\hline
\texttt{fr\_azo} & Number of azo groups \\
\hline
\texttt{fr\_barbitur} & Number of barbiturate groups \\
\hline
\texttt{fr\_benzene} & Number of benzene rings \\
\hline
\texttt{fr\_benzodiazepine} & Number of benzodiazepines with no additional fused rings \\
\hline
\texttt{fr\_bicyclic} & Bicyclic \\

\hline
\end{tabular}
\caption{List of Fragment Based Functional Groups tools (1/3)}
\label{tab:Fragment_Based_Functional_Groups}
\end{table*}

\begin{table*}[ht]
\centering
\begin{tabular}{|l|p{0.58\textwidth}|}
\hline
\textbf{Function} & \textbf{Description} \\
\hline
\texttt{fr\_diazo} & Number of diazo groups \\
\hline
\texttt{fr\_dihydropyridine} & Number of dihydropyridines \\
\hline
\texttt{fr\_epoxide} & Number of epoxide rings \\
\hline
\texttt{fr\_ester} & Number of esters \\
\hline
\texttt{fr\_ether} & Number of ether oxygens (including phenoxy) \\
\hline
\texttt{fr\_furan} & Number of furan rings \\
\hline
\texttt{fr\_guanido} & Number of guanidine groups \\
\hline
\texttt{fr\_halogen} & Number of halogens \\
\hline
\texttt{fr\_hdrzine} & Number of hydrazine groups \\
\hline
\texttt{fr\_hdrzone} & Number of hydrazone groups \\
\hline
\texttt{fr\_imidazole} & Number of imidazole rings \\
\hline
\texttt{fr\_imide} & Number of imide groups \\
\hline
\texttt{fr\_isocyan} & Number of isocyanates \\
\hline
\texttt{fr\_isothiocyan} & Number of isothiocyanates \\
\hline
\texttt{fr\_ketone} & Number of ketones \\
\hline
\texttt{fr\_ketone\_Topliss} & Number of ketones excluding diaryl, a,b-unsat. dienones, heteroatom on Calpha \\
\hline
\end{tabular}
\caption{List of Fragment Based Functional Groups tools (2/3)}
\label{tab:Fragment_Based_Functional_Groups_2}
\end{table*}
\begin{table*}[ht]
\centering
\begin{tabular}{|l|p{0.58\textwidth}|}
\hline
\textbf{Function} & \textbf{Description} \\
\hline
\texttt{fr\_lactam} & Number of beta lactams \\
\hline
\texttt{fr\_lactone} & Number of cyclic esters (lactones) \\
\hline
\texttt{fr\_methoxy} & Number of methoxy groups -OCH3 \\
\hline
\texttt{fr\_morpholine} & Number of morpholine rings \\
\hline
\texttt{fr\_nitrile} & Number of nitriles \\
\hline
\texttt{fr\_nitro} & Number of nitro groups \\
\hline
\texttt{fr\_nitro\_arom} & Number of nitro benzene ring substituents \\
\hline
\texttt{fr\_nitro\_arom\_nonortho} & Number of non-ortho nitro benzene ring substituents \\
\hline
\texttt{fr\_nitroso} & Number of nitroso groups, excluding NO2 \\
\hline
\texttt{fr\_oxazole} & Number of oxazole rings \\
\hline
\texttt{fr\_oxime} & Number of oxime groups \\
\hline
\texttt{fr\_para\_hydroxylation} & Number of para-hydroxylation sites \\
\hline
\texttt{fr\_phenol} & Number of phenols \\
\hline
\texttt{fr\_phenol\_noOrthoHbond} & Number of phenolic OH excluding ortho intramolecular Hbond substituents \\
\hline
\texttt{fr\_phos\_ester} & Number of phosphoric ester groups \\
\hline
\texttt{fr\_piperdine} & Number of piperdine rings \\
\hline
\texttt{fr\_piperzine} & Number of piperzine rings \\
\hline
\texttt{fr\_priamide} & Number of primary amides \\
\hline
\texttt{fr\_prisulfonamd} & Number of primary sulfonamides \\
\hline
\texttt{fr\_pyridine} & Number of pyridine rings \\
\hline
\texttt{fr\_quatN} & Number of quaternary nitrogens \\
\hline
\texttt{fr\_sulfide} & Number of thioether \\
\hline
\texttt{fr\_sulfonamd} & Number of sulfonamides \\
\hline
\texttt{fr\_sulfone} & Number of sulfone groups \\
\hline
\texttt{fr\_term\_acetylene} & Number of terminal acetylenes \\
\hline
\texttt{fr\_tetrazole} & Number of tetrazole rings \\
\hline
\texttt{fr\_thiazole} & Number of thiazole rings \\
\hline
\texttt{fr\_thiocyan} & Number of thiocyanates \\
\hline
\texttt{fr\_thiophene} & Number of thiophene rings \\
\hline
\texttt{fr\_unbrch\_alkane} & Number of unbranched alkanes of at least 4 members (excludes halogenated alkanes) \\
\hline
\texttt{fr\_urea} & Number of urea groups \\
\hline
\end{tabular}
\caption{List of Fragment Based Functional Groups tools (3/3)}
\label{tab:Fragment_Based_Functional_Groups_3}
\end{table*}

\begin{table*}[ht]
\centering
\begin{tabular}{|l|p{0.58\textwidth}|}
\hline
\textbf{Function} & \textbf{Description} \\
\hline
\texttt{get\_rdkit\_complexity} & Returns the Bertz molecular complexity index of the molecule. \\
\hline
\texttt{get\_rdkit\_number\_of\_atoms} & Returns the number of atoms in the molecule. \\
\hline
\texttt{get\_rdkit\_number\_of\_bonds} & Returns the number of bonds in the molecule. \\
\hline
\texttt{get\_rdkit\_rotatable\_bond\_count} & Returns the number of rotatable bonds in the molecule. \\
\hline
\texttt{get\_rdkit\_h\_bond\_donor\_count} & Returns the number of hydrogen bond donors in the molecule. \\
\hline
\texttt{get\_rdkit\_h\_bond\_acceptor\_count} & Returns the number of hydrogen bond acceptors in the molecule. \\
\hline
\texttt{get\_rdkit\_molecular\_formula} & Returns the molecular formula of the molecule. \\
\hline
\texttt{get\_rdkit\_canonical\_smiles} & Returns the canonical SMILES of the molecule. \\
\hline
\texttt{get\_rdkit\_inchi} & Returns the InChI string of the molecule. \\
\hline
\end{tabular}
\caption{List of Identifiers and Representations tools}
\label{tab:Identifiers_and_Representations}
\end{table*}

\begin{table*}[ht]
\centering
\begin{tabular}{|l|p{0.58\textwidth}|}
\hline
\textbf{Function} & \textbf{Description} \\
\hline
\texttt{smi2mol\_with\_errors} & Attempts to parse SMILES and returns validation status with error/warning messages. \\
\hline
\texttt{calcmolformula} & Returns the molecule 2019s formula \\
\hline
\texttt{calccrippendescriptors} & Returns a 2-tuple with the Wildman-Crippen logp,mr values  \\
\hline
\texttt{calcfractioncsp3} & Returns the fraction of C atoms that are SP3 hybridized  \\
\hline
\texttt{calckappa1} &  Calculates the first Kier shape index, reflecting molecular linearity based on atom and bond counts.\\
\hline
\texttt{calckappa2} & Computes the second Kier shape index, indicating molecular cyclicity and branching. \\
\hline
\texttt{calckappa3} &  Computes the third Kier shape index, sensitive to molecular flexibility and complex ring structures.\\
\hline
\texttt{calclabuteasa} & Returns the Labute ASA value for a molecule  \\
\hline
\texttt{calcpbf} & Returns the PBF (plane of best fit) descriptor   \\
\hline
\texttt{calcphi} &  Estimates the molecular flexibility index based on the number of rotatable bonds and ring structures.\\
\hline
\texttt{getconnectivityinvariants} & Returns connectivity invariants (ECFP-like) for a molecule.  \\
\hline
\texttt{getfeatureinvariants} & Returns feature invariants (FCFP-like) for a molecule.  \\
\hline

\texttt{mqns\_} & Computes Molecular Quantum Numbers, a 42-dimensional vector of counts for various atom types, bonds, and topological features.\\
\hline
\texttt{peoe\_vsa\_} &   Computes descriptors combining partial charges (Gasteiger) with van der Waals surface areas in defined bins.\\
\hline
\texttt{smr\_vsa\_} &  Calculates descriptors combining molar refractivity contributions with surface areas in predefined ranges.\\
\hline
\texttt{slogp\_vsa\_} & Computes descriptors by combining atomic logP contributions (Wildman-Crippen) with van der Waals surface areas. \\
\hline
\end{tabular}
\caption{List of Other Descriptors tools}
\label{tab:Other_Descriptors}
\end{table*}
\begin{table*}[ht]

\centering
\begin{tabular}{|l|p{0.58\textwidth}|}
\hline
\textbf{Function} & \textbf{Description} \\
\hline
\texttt{get\_center} & Computes the geometric center of a conformer generated from the input SMILES. \\
\hline
\texttt{get\_shape\_moments} & Calculates NPR1 and NPR2 shape descriptors from a generated conformer. \\
\hline
\texttt{refine\_conformers} & Refines 3D conformers based on energy and RMSD thresholds. \\
\hline
\texttt{get\_conformer\_energies} & Returns the energies of multiple conformers generated from the input molecule. \\
\hline
\texttt{calcnumaliphaticcarbocycles} & Returns the number of aliphatic (containing at least one non-aromatic bond) carbocycles for a molecule \\
\hline
\texttt{calcnumaliphaticheterocycles} & Returns the number of aliphatic (containing at least one non-aromatic bond) heterocycles for a molecule \\
\hline
\texttt{calcnumaliphaticrings} & Returns the number of aliphatic (containing at least one non-aromatic bond) rings for a molecule\\
\hline
\texttt{calcnumamidebonds} & Returns the number of amide bonds in a molecule \\
\hline
\texttt{calcnumaromaticcarbocycles} & Returns the number of aromatic carbocycles for a molecule\\
\hline
\texttt{calcnumaromaticheterocycles} & Returns the number of aromatic heterocycles for a molecule\\
\hline
\texttt{calcnumaromaticrings} & Returns the number of aromatic rings for a molecule  \\
\hline
\texttt{calcnumatomstereocenters} & Returns the total number of atomic stereocenters (specified and unspecified)  \\
\hline
\texttt{calcnumatoms} & Returns the total number of atoms for a molecule \\
\hline
\texttt{calcnumhba} & Returns the number of H-bond acceptors for a molecule \\
\hline
\texttt{calcnumhbd} & Returns the number of H-bond donors for a molecule \\
\hline
\texttt{calcnumheavyatoms} & Returns the number of heavy atoms for a molecule \\
\hline
\texttt{calcnumheteroatoms} & Returns the number of heteroatoms for a molecule  \\
\hline
\texttt{calcnumheterocycles} & Returns the number of heterocycles for a molecule  \\
\hline
\texttt{calcnumlipinskihba} & Returns the number of Lipinski H-bond acceptors for a molecule  \\
\hline
\texttt{calcnumlipinskihbd} & Returns the number of Lipinski H-bond donors for a molecule \\
\hline
\texttt{calcnumrings} & Returns the number of rings for a molecule \\
\hline
\texttt{calcnumrotatablebonds} & Returns the number of rotatable bonds for a molecule. strict = NumRotatableBondsOptions.NonStrict - Simple rotatable bond definition. \\
\hline
\texttt{calcnumsaturatedcarbocycles} & Returns the number of saturated carbocycles for a molecule  \\
\hline
\texttt{calcnumsaturatedheterocycles} & Returns the number of saturated heterocycles for a molecule  \\
\hline
\texttt{calcnumsaturatedrings} & Returns the number of saturated rings for a molecule  \\
\hline
\texttt{calcnumunspecifiedatomstereocenters} & Returns the number of unspecified atomic stereocenters \\
\hline
\texttt{calcoxidationnumbers} & Adds the oxidation number/state to the atoms of a molecule as property OxidationNumber on each atom. Use Pauling electronegativities. This is experimental code, still under development.  \\
\hline
\end{tabular}
\caption{List of Structural Descriptors tools}
\label{tab:Structural_Descriptors}
\end{table*}
    
List of tools are provided by categories. Tools of electronic and topological descriptors are provided at \Cref{tab:Electronic_Topological_Descriptors}, fragment based functional groups at \Cref{tab:Fragment_Based_Functional_Groups}, \Cref{tab:Fragment_Based_Functional_Groups_2}, and \Cref{tab:Fragment_Based_Functional_Groups_3}, identifiers and representations at \Cref{tab:Identifiers_and_Representations}, structural descriptors at \Cref{tab:Structural_Descriptors}, and miscellaneous descriptors at \Cref{tab:Other_Descriptors}.

\newpage
\clearpage
\section{Prompts}\label{appdx:prompts}
\subsection{Prompt for analyst}
\begin{GrayBox}
    You are a professional AI chemistry assistant specialized in resolving [category name] using RDKit tools.\\
    Your job is to identify how to retrieve standardized molecular information such as CIDs, InChI, and canonical SMILES for downstream processing.\\

    Follow this structured reasoning process step-by-step:\\

    Step 1. Analyze the molecule design condition which is the goal of the task.\\
    Step 2. Parse list of all valid SMILES strings mentioned anywhere in the user prompt and output them in the provided JSON format.\\
    Step 3. Based on your chemical knowledge, explain why standardizing identifiers and resolving canonical formats might be important for this task.\\
        - E.g., checking uniqueness, linking to external data, verifying molecular identity.\\
    Step 4. Choose as many tools as necessary from the identifier toolset that help you access consistent molecular representations or external references.\\
    Step 5. Output your final answer in the provided JSON format.\\

    This is a molecule design condition of the [task name] task: [task description]
                
Now output the tools to use by using the following JSON format.
Take a deep breath and think carefully before writing your answer.
```json
$\lbrace\lbrace$ \\
\quad"parsed\_smiles": [\\
\quad$\lbrace\lbrace$\\
\qquad"smiles": "Parsed SMILES string",\\
    $\rbrace\rbrace$  ,\\
    ...\\
  ],\\
      "tools\_to\_use": [\\
    $\lbrace\lbrace$\\
      "tool\_name": "fr\_Ar\_OH",\\
      "purpose": "Detect aromatic hydroxyl groups, similar to those in albuterol."\\
    $\rbrace\rbrace$,\\
    ...\\
  ]\\
$\rbrace\rbrace$
\end{GrayBox}

\newpage
\clearpage
\subsection{Prompt for scientist}
\begin{GrayBox} You are a skilled chemist.\\

Your task is to design a SMILES string for a molecule that satisfies the following condition: [task description]\\

Functional groups and molecule tool analysis results of task related molecules: [result of tool analysis]\\

You are provided with:\\
- Top-100 example molecules with high relevance to the task, listed below. You may use these as inspiration, but YOU MUST NOT COPY THEM EXACTLY.\\
- A list of previously generated SMILES, which YOU MUST NOT REPEAT.\\

Top-100 Relevant SMILES Examples (SMILES, score)\\
YOU MUST FAITHFULLY REFER TO THESE EXAMPLES WHEN DESIGNING YOUR MOLECULE. BUT DO NOT COPY THEM EXACTLY: \\

[top100 SMLIES]\\

You must return your response in the following json format.
The text inside each key explains what kind of answer is expected — it is a guideline, not the answer.\\

DO NOT repeat the example text or instructions. 
Instead, write your own scientifically reasoned content based on the task.\\

Use the following format.
Take a deep breath and think carefully before writing your answer.\\

```json\\
$\lbrace \lbrace$\\
  "step1": "List of the target's critical structural/property features (e.g., 'Target: phenyl ring, $\beta$-hydroxyamine, catechol-like substitution'). If property-based, specify requirements (e.g., "logP > 3: add hydrophobic groups").",\\
  "step2": "Propose modifications or scaffolds to meet the condition (e.g., 'Replace catechol with 3-hydroxy-4-pyridone').\\ Justify each change chemically (e.g., "Maintains H-bonding but improves metabolic stability").",\\
  "step3": "Describe the full structure of your designed molecule in natural language before writing the SMILES. (e.g., "A tert-butyl group attached to the amine (–NH–C(CH₃)₃) to mimic target's bulky substituent.")",\\
  "smiles": "Your valid SMILES string here"\\
$\rbrace\rbrace$
\end{GrayBox}

\newpage
\clearpage
\subsection{Prompts for scientist with feedback}
\begin{GrayBox}
YOU MUST NOT REPEAT ANY OF THE PREVIOUSLY GENERATED SMILES:
[smiles\_history]\\
Task: Take [verifier/reviewer]'s feedback actively and design a SMILES string for a molecule that satisfies the condition:\\

Condition for molecule design:   

[task description]\\

Functional groups and molecule tool analysis results of task related molecules:

[target functional groups]\\

Top-100 Relevant SMILES Examples (SMILES, score)\\
YOU MUST FAITHFULLY REFER TO THESE EXAMPLES WHEN DESIGNING YOUR MOLECULE. BUT DO NOT COPY THEM EXACTLY:

[topk smiles]\\

You will be provided with:\\
1. Previous SMILES string\\
2. Task score (0–1)\\
3. Detected functional groups in your previous molecule \\

--- MOLECULE SMILES TO IMPROVE ---\\
MOLECULE SMILES: [previous smiles]\\
- Task score: [score] (0–1)\\
- Functional groups detected:

[functional groups]\\

--- YOUR PREVIOUS THOUGHT AND REVIEWER'S FEEDBACK ---\\
Step1: List Key Features\\
Your previous thought process:

[scientist step1 reasoning]\\
Accordingly, reviewer's feedback is:

[verifier/reviewer step1 feedback]\\
    
Step2: Design Strategy:\\
Your previous thought process:

[scientist step2 think]\\
Accordingly, reviewer's feedback is:

[verifier/reviewer step2 feedback]\\

Step 3: Construct the Molecule:
Your previous thought process:

[verifier/scientist step3 think]\\
Accordingly, reviewer's feedback is:

[verifier/reviewer step3 feedback]\\

Now based on your previous thoughts and the reviewer's feedback, you need to improve your design.
You must return your response in the following json format.\\
The text inside each key explains what kind of answer is expected — it is a guideline, not the answer.\\

\end{GrayBox}
\newpage
\clearpage
\begin{GrayBox}
    
DO NOT repeat the example text or instructions.  \\
Instead, write your own scientifically reasoned content based on the task.\\

Use the following format.\\
Take a deep breath and think carefully before writing your answer.\\
```json\\
$\lbrace\lbrace$\\
  "step1": "List of the target's critical structural/property features (e.g., 'Target: phenyl ring, $\beta$-hydroxyamine, catechol-like substitution'). If property-based, specify requirements (e.g., "logP > 3: add hydrophobic groups").",\\
  "step2": "Propose modifications or scaffolds to meet the condition (e.g., 'Replace catechol with 3-hydroxy-4-pyridone'). Justify each change chemically (e.g., "Maintains H-bonding but improves metabolic stability").",\\
  "step3": "Describe the full structure of your designed molecule in natural language before writing the SMILES. (e.g., "A tert-butyl group attached to the amine (–NH–C(CH₃)₃) to mimic target's bulky substituent.")",\\
  "smiles": "Your valid SMILES string here"\\
$\rbrace\rbrace$\\
```
\end{GrayBox}
\newpage
\clearpage
\subsection{Prompts for verifier}

\begin{GrayBox}
    You are a meticulous double-checker LLM. Your task is to verify whether each step of the scientist’s reasoning is chemically valid and faithfully and logically reflected in the final SMILES string. \\
    You will be given: \\
- A user prompt describing the target objective,\\
- The scientist's reasoning broken into Step1 through Step3,\\
- The SMILES string proposed by the scientist.\\
Evaluate each step independently, comparing the described logic to the molecular structure in the SMILES.

Provide a reasoning assessment for each step.
=== SCIENTIST'S TASK === \\
If any step is inconsistent, mark "Consistency" as "Inconsistent" and provide specific suggestions for improvement.\\

[task description]\\

Functional groups and molecule tool analysis results of task related molecules:

[target functional groups]\\

=== SCIENTIST'S THINKING ===\\ 
Step1: [thinking['step1']]\\
Step2: [thinking['step2']]\\
Step3: [thinking['step3']]\\

=== SCIENTIST'S SMILES === \\
- SMILES: [smiles]\\
- Detected functional groups and molecule tool analysis results:\\

[functional groups]

You must return your response in the following json format.\\
The text inside each key explains what kind of answer is expected — it is a guideline, not the answer.\\

DO NOT repeat the example text or instructions.  \\
Instead, write your own scientifically reasoned content based on the task.\\

Use the following format.\\
Take a deep breath and think carefully before writing your answer. \\
```json
$\lbrace\lbrace$\\
  "step1": "Your analysis of whether scientist's Step1 thinking is chemically valid and  reflected in the SMILES.",\\
  "step2": "Your analysis of whether scientist's Step2 thinking is chemically valid and  reflected in the SMILES.",\\
  "step3": "Your analysis of whether scientist's Step3 thinking is chemically valid and reflected in the SMILES.",\\
  "consistency": "Consistent" or "Inconsistent",\\
$\rbrace\rbrace$\\
```
\end{GrayBox}

\newpage
\clearpage
\subsection{Prompts for reviewer}
\begin{GrayBox}
You are a rigorous chemistry reviewer.\\
Evaluate the Scientist LLM's reasoning steps and final SMILES molecule for:\\
- Validity\\
- Chemical soundness\\
- Adherence to the design condition:\\

Scientist LLM's task:

[task description]\\

Be constructive: Provide fixes for issues (e.g., "Replace C=O=C with O=C=O for carbon dioxide").\\

You are provided with:\\
- Scientist's thinking\\
- Scientist-generated SMILES\\
- Task score\\
- Detected functional groups in the generated molecule\\

--- SCIENTIST'S STEP-WISE THINKING --- \\
Step 1: [scientist step1 reasoning] \\

Step 2: [scientist step2 reasoning] \\

Step 3: [scientist step3 reasoning] \\

--- SCIENTIST-MOLECULE SMILES ---\\
SMILES: [scientist proposed SMILES]\\
- Task score: [score] (range: 0 to 1)\\
- Detected functional groups and molecule tool analysis results:

[functional groups]\\

You must return your response in the following json format.\\
The text inside each key explains what kind of answer is expected — it is a guideline, not the answer.\\

DO NOT repeat the example text or instructions.  \\
Instead, write your own scientifically reasoned content based on the task.\\

Use the following format.\\
Take a deep breath and think carefully before writing your answer.\\
```json\\
$\lbrace\lbrace$\\
  "step1": "List accurate features and functional groups identified.
  Mention any critical features and functional groups that were missed or misinterpreted.",\\
  "step2": "Evaluate if the proposed design strategy aligns with the structural and functional similarity goal.\\
  Comment on whether the design aligns with the initial objectives.
  Suggest improvements or alternatives if needed.",\\

```
\end{GrayBox}
\newpage
\clearpage
\begin{GrayBox}
  "step3": "Review the structural construction and positional assignments.
  Check for missing elements or mismatches in reasoning. (e.g., "Claimed 'para hydroxyl' but SMILES places it meta")",\\
$\rbrace\rbrace$ \\
\end{GrayBox}
\newpage
\clearpage
\section{Task description}\label{appdx:task_description}
In this section, we describe the 23 tasks of practical molecular benchmark \cite{gao2022sample}. For more details about the task and oracle, refer to Therapeutics Data Commons \citep[TDC]{huang2021therapeutics} document: \url{https://tdc.readthedocs.io/en/main/_modules/tdc/chem_utils/oracle/oracle.html}.

% \newline

\textbf{1. albuterol\_similarity}

Design a molecule similar to albuterol while preserving key functional groups.  
\newline

\textbf{2. amlodipine\_mpo  }

Generate molecules similar to amlodipine with good drug-like properties (e.g., 3-ring topology).  
\newline

\textbf{3. celecoxib\_rediscovery  }

Recreate the anti-inflammatory drug celecoxib.  
\newline

\textbf{4. deco\_hop  }

Modify the decorations of a molecule while preserving a fixed scaffold.  
Avoid forbidden substructures and stay below similarity cap.
\newline

\textbf{5. drd2  }

Generate molecules predicted to strongly bind to the dopamine D2 receptor using a predictive model.
\newline

\textbf{6. fexofenadine\_mpo  }

Create molecules structurally similar to fexofenadine with TPSA $\approx$ 90 and logP $\approx$ 4.  
\newline

\textbf{7. gsk3b  }

Design molecules predicted to have high binding affinity for the GSK3$\beta$ protein.
\newline

\textbf{8. isomers\_c7h8n2o2  }

Generate any molecule that is an exact isomer of C\textsubscript{7}H\textsubscript{8}N\textsubscript{2}O\textsubscript{2}.  
Must match the molecular formula exactly.
\newline

\textbf{9. isomers\_c9h10n2o2pf2cl  }

Generate an exact isomer of C\textsubscript{9}H\textsubscript{10}N\textsubscript{2}O\textsubscript{2}PF\textsubscript{2}Cl.
\newline

\textbf{10. jnk3  }

Design molecules with high predicted inhibitory activity against the JNK3 protein.
\newline

\vspace{5in}
\textbf{11. median1}  

Find a molecule similar to both camphor and menthol.  
\newline

\textbf{12. median2 } 

Design a molecule similar to both tadalafil and sildenafil.
\newline

\textbf{13. mestranol\_similarity  }

Generate molecules similar to the hormone mestranol, preserving the core scaffold.
\newline

\textbf{14. osimertinib\_mpo  }

Create osimertinib-like molecules with low logP ($\approx$1) and TPSA $\approx$ 100. 
\newline

\textbf{15. perindopril\_mpo  }

Design perindopril-like molecules.  
\newline

\textbf{16. qed  }

Maximize a quantitative estimate of drug-likeness (QED) score.
\newline
\textbf{17. ranolazine\_mpo  }

Create ranolazine-like molecules with TPSA $\approx$ 95 and logP $\approx$ 7. 
\newline

\textbf{18. scaffold\_hop  }

Replace the molecular scaffold while keeping key functional groups unchanged.
\newline

\textbf{19. sitagliptin\_mpo  }

Design sitagliptin-like molecules matching formula C\textsubscript{16}H\textsubscript{15}F\textsubscript{6}N\textsubscript{5}O.
\newline

\textbf{20. thiothixene\_rediscovery  }

Reproduce the structure of thiothixene.  
\newline

\textbf{21. troglitazone\_rediscovery  }

Reconstruct the diabetes drug troglitazone.  
\newline

\textbf{22. valsartan\_smarts  }

Generate molecules containing the substructure SMARTS with logP $\approx$ 2.0 and TPSA $\approx$ 95.
\newline

\textbf{23. zaleplon\_mpo}

Design zaleplon-like molecules with formula C\textsubscript{19}H\textsubscript{17}N\textsubscript{3}O\textsubscript{2}. 
\newpage
\clearpage
\section{PMO-1K experiment result}\label{appdx:pmo_full}

\begin{table*}[t] \label{tab:appdx_pmo_table}
\centering
\resizebox{\textwidth}{!}{
\begin{tabular}{lccccccccccc}
\hline
\textbf{Task} & \textbf{GP BO} & \textbf{REINVENT} & \textbf{LICO}\textit{-L} & \textbf{Genetic GFN} & \textbf{Graph GA} & \textbf{Aug. Mem.} & \textbf{MOLLEO}\textit{-B} & \textbf{MOLLEO}\textit{-D}* & \textbf{Ours}\textit{-D}* \\
\hline
albuterol\_similarity & $0.636 \pm 0.106$ & $0.496 \pm 0.020$ & $0.656 \pm 0.125$ & $0.664 \pm 0.054$ & $0.583 \pm 0.065$ & $0.557 \pm 0.048$ & $\underline{{0.886 \pm 0.023}}$ & $0.883 \pm 0.001$ & $\mathbf{{0.998 \pm 0.000}}$ \\
amlodipine\_mpo & $0.519 \pm 0.014$ & $0.472 \pm 0.008$ & $0.541 \pm 0.026$ & $0.534 \pm 0.019$ & $0.501 \pm 0.016$ & $0.489 \pm 0.009$ & $\underline{{0.637 \pm 0.023}}$ & $0.540 \pm 0.072$ & $\mathbf{{0.647 \pm 0.010}}$ \\
celecoxib\_rediscovery & $0.411 \pm 0.046$ & $0.370 \pm 0.029$ & $0.447 \pm 0.073$ & $0.447 \pm 0.028$ & $0.424 \pm 0.049$ & $0.385 \pm 0.027$ & $0.402 \pm 0.003$ & $\underline{{0.512 \pm 0.119}}$ & $\mathbf{{0.867 \pm 0.007}}$ \\
deco\_hop & $0.593 \pm 0.018$ & $0.572 \pm 0.006$ & $0.596 \pm 0.010$ & $\underline{{0.604\pm 0.017}} $ & $0.581 \pm 0.006$ & $0.579 \pm 0.010$ & $0.588 \pm 0.007$ & $0.574 \pm 0.001$ & $\mathbf{{0.842 \pm 0.077}}$ \\
drd2 & $0.857 \pm 0.080$ & $0.775 \pm 0.086$ & $\underline{{0.859 \pm 0.066}}$ & $0.809 \pm 0.045$ & $0.833 \pm 0.065$ & $0.795 \pm 0.024$ & $\mathbf{{0.910 \pm 0.017}}$ & $0.812 \pm 0.027$ & $0.756\pm0410$ \\
fexofenadine\_mpo & $\underline{{0.707 \pm 0.021}}$ & $0.650 \pm 0.007$ & $0.700 \pm 0.023$ & $0.682 \pm 0.021$ & $0.666 \pm 0.009$ & $0.679 \pm 0.021$ & $0.674 \pm 0.002$ & $0.680 \pm 0.007$ & $\mathbf{{0.883 \pm 0.02}}$ \\
gsk3b & $0.611 \pm 0.059$ & $0.589 \pm 0.063$ & $\underline{{0.617 \pm 0.063}}$ & $\mathbf{{0.637 \pm 0.018}}$ & $0.523 \pm 0.047$ & $0.539 \pm 0.097$ & $0.397 \pm 0.013$ & $0.496 \pm 0.073$ & $0.308\pm0.009$ \\
isomers\_c7h8n2o2 & $0.545 \pm 0.158$ & $0.725 \pm 0.064$ & $0.779 \pm 0.099$ & $0.738 \pm 0.039$ & $0.735 \pm 0.112$ & $0.661 \pm 0.039$ & $0.737 \pm 0.043$ & $\underline{{0.850 \pm 0.009}}$ & $\mathbf{{0.986 \pm 0.015}}$ \\
isomers\_c9h10n2o2pf2cl & $0.599 \pm 0.059$ & $0.630 \pm 0.032$ & $0.672 \pm 0.075$ & $0.656 \pm 0.075$ & $0.630 \pm 0.086$ & $0.596 \pm 0.066$ & $0.635 \pm 0.017$ & $\underline{{0.832 \pm 0.007}}$ & $\mathbf{{0.914 \pm 0.031}}$ \\
jnk3 & $\underline{{0.346 \pm 0.067}}$ & $0.315 \pm 0.042$ & $0.336 \pm 0.051$ & $\mathbf{{0.409 \pm 0.165}}$ & $0.301 \pm 0.071$ & $0.294 \pm 0.110$ & $0.186 \pm 0.076$ & $0.342 \pm 0.044$ & $0.125\pm0.020$ \\
median1 & $0.213 \pm 0.020$ & $0.205 \pm 0.014$ & $0.217 \pm 0.019$ & $0.219 \pm 0.008$ & $0.208 \pm 0.015$ & $0.219 \pm 0.014$ & $\underline{{0.236 \pm 0.021}}$ & $0.193 \pm 0.005$ & $\mathbf{{0.321 \pm 0.029}}$ \\
median2 & $0.203 \pm 0.009$ & $0.188 \pm 0.010$ & $0.193 \pm 0.009$ & $\underline{{0.204 \pm 0.011}}$ & $0.181 \pm 0.009$ & $0.184 \pm 0.010$ & $0.191 \pm 0.009$ & $0.197 \pm 0.023$ & $\mathbf{{0.322 \pm 0.024}}$ \\
mestranol\_similarity & $0.427 \pm 0.025$ & $0.379 \pm 0.026$ & $0.423 \pm 0.016$ & $0.414 \pm 0.022$ & $0.362 \pm 0.017$ & $0.393 \pm 0.021$ & $0.399 \pm 0.020$ & $\underline{{0.630 \pm 0.171}}$ & $\mathbf{{0.996 \pm 0.001}}$ \\
osimertinib\_mpo & $0.766 \pm 0.006$ & $0.737 \pm 0.007$ & $0.759 \pm 0.008$ & $0.763 \pm 0.008$ & $0.751 \pm 0.005$ & $0.761 \pm 0.006$ & $\underline{{0.779 \pm 0.006}}$ & $0.753 \pm 0.018$ & $\mathbf{{0.796 \pm 0.005}}$ \\
perindopril\_mpo & $0.458 \pm 0.019$ & $0.404 \pm 0.008$ & $0.473 \pm 0.009$ & $0.462 \pm 0.033$ & $0.435 \pm 0.016$ & $0.422 \pm 0.013$ & $\mathbf{{0.655 \pm 0.054}}$ & $0.422 \pm 0.006$ & $\underline{{0.542 \pm 0.027}}$ \\
qed & $0.912 \pm 0.010$ & $0.921 \pm 0.002$ & $\underline{{0.925 \pm 0.005}}$ & $\mathbf{{0.928 \pm 0.002}}$ & $0.914 \pm 0.007$ & $0.923 \pm 0.002$ & $0.919 \pm 0.006$ & $\mathbf{{0.928 \pm 0.006}}$ & $0.903\pm0.003$ \\
ranolazine\_mpo & $\mathbf{{0.701 \pm 0.023}}$ & $0.574 \pm 0.044$ & $\underline{{0.687 \pm 0.029}}$ & $0.623 \pm 0.022$ & $0.620 \pm 0.014$ & $0.614 \pm 0.033$ & $0.640 \pm 0.000$ & $0.516 \pm 0.024$ & $0.233\pm0.018$ \\
scaffold\_hop & $0.478 \pm 0.009$ & $0.447 \pm 0.010$ & $0.480 \pm 0.008$ & $\underline{{0.485 \pm 0.015}}$ & $0.461 \pm 0.008$ & $0.460 \pm 0.010$ & $0.473 \pm 0.000$ & $0.464 \pm 0.002$ & $\mathbf{{0.646 \pm 0.055}}$ \\
sitagliptin\_mpo & $0.232 \pm 0.083$ & $0.261 \pm 0.026$ & $\underline{{0.315 \pm 0.097}}$ & $0.227 \pm 0.041$ & $0.229 \pm 0.053$ & $0.245 \pm 0.030$ & $0.193 \pm 0.073$ & $\mathbf{{0.328 \pm 0.091}}$ & $0.067\pm0.006$ \\
thiothixene\_rediscovery & $0.351 \pm 0.033$ & $0.311 \pm 0.021$ & $0.343 \pm 0.035$ & $0.377 \pm 0.015$ & $0.322 \pm 0.023$ & $0.336 \pm 0.073$ & $0.416 \pm 0.075$ & $\underline{{0.478 \pm 0.028}}$ & $\mathbf{{0.719 \pm 0.001}}$ \\
troglitazone\_rediscovery & $0.313 \pm 0.018$ & $0.246 \pm 0.009$ & $0.292 \pm 0.028$ & $0.277 \pm 0.015$ & $0.267 \pm 0.015$ & $0.262 \pm 0.012$ & $0.302 \pm 0.022$ & $\underline{{0.387 \pm 0.013}}$ & $\mathbf{{0.841 \pm 0.042}}$ \\
valsartan\_smarts & $0.000 \pm 0.000$ & $0.000 \pm 0.000$ & $0.000 \pm 0.000$ & $0.000 \pm 0.000$ & $0.000 \pm 0.000$ & $0.000 \pm 0.000$ & $0.000 \pm 0.000$ & $0.000 \pm 0.000$ & $0.000 \pm 0.000$ \\
zaleplon\_mpo & $0.392 \pm 0.034$ & $0.406 \pm 0.017$ & $0.404 \pm 0.022$ & $0.400 \pm 0.014$ & $0.374 \pm 0.024$ & $\underline{{0.415 \pm 0.013}}$ & $0.392 \pm 0.003$ & $0.409 \pm 0.005$ & $\mathbf{{0.625 \pm 0.046}}$ \\

\hline
\textbf{Sum of scores (↑)} & $11.27$ & $10.68$ & $11.71$ & $11.56$ & $10.90$ & $10.81$ & $11.65$ & $\underline{{12.23}}$ & $\mathbf{{15.42}}$ \\
\hline
\end{tabular}
}
\caption{\textbf{Detailed results of PMO-1K benchmark.} Tasks are assessed using AUC top-10 with mean $\pm$ standard deviation. Results with (*) are evaluated from 3 independent runs while the others are assessed from 5 independent runs. We mark the best result in \textbf{bold} and the second-best are \underline{underlined}  for each task.
}
\label{tab:appdx_pmo_table}
\end{table*}

We provide the full PMO-1K experiment result in \Cref{tab:appdx_pmo_table}.
\section{ZINC 250K statistics}
\begin{table*}[ht]
\centering
\begin{tabular}{lcccc}
\hline
\textbf{Oracle} & \textbf{Min} & \textbf{Max} & \textbf{Mean} & \textbf{Std} \\
\hline
albuterol\_similarity & 0.053 & 0.667 & 0.251 & 0.062 \\
amlodipine\_mpo & 0.000 & 0.686 & 0.214 & 0.144 \\
celecoxib\_rediscovery & 0.000 & 0.447 & 0.142 & 0.060 \\
deco\_hop & 0.291 & 0.878 & 0.768 & 0.048 \\
drd2 & 0.000 & 0.987 & 0.009 & 0.038 \\
fexofenadine\_mpo & 0.000 & 0.756 & 0.232 & 0.206 \\
gsk3b & 0.000 & 0.990 & 0.030 & 0.045 \\
isomers\_c7h8n2o2 & 0.000 & 1.000 & 0.004 & 0.037 \\
isomers\_c9h10n2o2pf2cl & 0.000 & 0.869 & 0.018 & 0.071 \\
jnk3 & 0.000 & 0.680 & 0.016 & 0.026 \\
median1 & 0.000 & 0.324 & 0.066 & 0.037 \\
median2 & 0.000 & 0.291 & 0.108 & 0.027 \\
mestranol\_similarity & 0.004 & 0.886 & 0.170 & 0.059 \\
osimertinib\_mpo & 0.000 & 0.829 & 0.179 & 0.209 \\
perindopril\_mpo & 0.000 & 0.560 & 0.176 & 0.113 \\
qed & 0.117 & 0.948 & 0.732 & 0.139 \\
ranolazine\_mpo & 0.000 & 0.586 & 0.059 & 0.069 \\
scaffold\_hop & 0.176 & 0.526 & 0.373 & 0.026 \\
sitagliptin\_mpo & 0.000 & 0.479 & 0.012 & 0.035 \\
thiothixene\_rediscovery & 0.000 & 0.408 & 0.162 & 0.047 \\
troglitazon\_rediscovery & 0.000 & 0.391 & 0.135 & 0.035 \\
valsartan\_smarts & 0.000 & 0.320 & 0.000 & 0.001 \\
zaleplon\_mpo & 0.000 & 0.545 & 0.072 & 0.100 \\
\hline
\end{tabular}
\caption{Data statistics of ZINC 250k that we retrieved for each oracle.}
\label{tab:score_statistics}
\end{table*}

We provide the data statistics of ZINC250K~\cite{sterling2015zinc} that we used in our setting at \Cref{tab:score_statistics}.
\section{Usage of AI assistants}\label{appx: ai_assistant}

We used AI writing assistants (e.g., ChatGPT) to improve the clarity, grammar, and style of the manuscript during the writing process. These tools were employed strictly for language refinement and did not contribute to the development of ideas, methods, or analysis. All scientific contributions and experimental results are the original work of the authors.

\section{Scientific Artifacts}\label{appx: artifact}

\paragraph{The License for artifacts.} 

We used dataset and tools accordingly with their respective licenses. In detail, We use open-source ZINC250K dataset~\citep{sterling2015zinc} and the publicly available RDKIt tools~\citep{landrum2013rdkit}. We provide our source code at \url{https://anonymous.4open.science/r/mt_mol-0448} for reproducibility with an appropriate open-source license.

\paragraph{Artifact use consistency with intended use.} 
We used dataset and tools in line of their intended use. Specifically, ZINC250K~\citep{sterling2015zinc} incorporates molecule with property scores for molecular optimization task which aligns with goal of our study. Also, RDKit tools are used to analyze the chemical properties of the given molecule which is used in our study.

\end{document}